\documentclass[conference]{IEEEtran}

  	\usepackage[pdftex]{graphicx}
  	\graphicspath{{../pdf/}{../jpeg/}}
	\DeclareGraphicsExtensions{.pdf,.jpeg,.png}

	\usepackage[cmex10]{amsmath}
	\usepackage{mathabx}
	\usepackage{algorithmic}
	\usepackage{array}
	\usepackage{mdwmath}
	\usepackage{mdwtab}
	\usepackage{eqparbox}
	\usepackage{url}
\usepackage{pifont}
\usepackage{multirow}
\usepackage[most]{tcolorbox}
\usepackage{placeins} 

\usepackage{todonotes}

\begin{document}

\title{\LARGE Enhancing Object Detection Performance for Small Objects through Synthetic Data Generation and Proportional Class-Balancing Technique: A Comparative Study in Industrial Scenarios}


 \author{\authorblockN{Jibinraj Antony\authorrefmark{2}\authorrefmark{3}, Vinit Hegiste\authorrefmark{1}, Ali Nazeri\authorrefmark{2}, Hooman Tavakoli\authorrefmark{2}, Snehal Walunj\authorrefmark{2}, \\ Christiane Plociennik\authorrefmark{2}, Martin Ruskowski \authorrefmark{2}\authorrefmark{1}}
 \authorblockA{\authorrefmark{2}German Research Center for Artificial Intelligence (DFKI), Kaiserslautern, Germany}
 \authorblockA{\authorrefmark{1}Rheinland-Pfälzische Technische Universität (RPTU) Kaiserslautern-Landau, Kaiserslautern, Germany}
 \authorblockA{\authorrefmark{3}\textit{Corresponding Author. Tel.: +49-176 6290 8490 ; E-mail : jibinraj.antony@dfki.de}\\ **\textit{The first five Authors contributed to the Research equally.}}}

\maketitle

\begin{abstract}
Object Detection (OD) has proven to be a significant computer vision method in extracting localized class information and has multiple applications in the industry. Although many of the state-of-the-art OD models perform well on medium and large sized objects, they seem to under perform on small objects. In most of the industrial use cases, it is difficult to collect and annotate data for small objects, as it is time-consuming and prone to human errors. Additionally, those datasets are likely to be unbalanced and often results in inefficient model convergence. To tackle this challenge, this study presents a novel approach that injects additional data points to improve the performance of the OD models. Using synthetic data generation, the difficulties in data collection and annotations for small object data points can be minimized and a balanced distribution of dataset can be created. This paper discusses the effects of a simple proportional class-balancing technique, to enable better anchor matching of the OD models. A comparison has been made on the performances of the state-of-the-art OD models: YOLOv5, YOLOv7 and SSD, for combinations of real and synthetic datasets within an industrial use case.
\end{abstract}

\IEEEoverridecommandlockouts

\IEEEpeerreviewmaketitle


\section{Introduction}\label{intro}

The birth of Convolution Neural Networks (CNN) set down an important milestone in the performance improvement of image classification challenges, where networks like AlexNet \cite{AlexNet} and their successors exceed human level performances. As the researches in CNN advanced, so did the academic interest in the localization of the specific objects, where networks like OverFeat \cite{sermanet2013overfeat} made their first success with a deep CNN architecture. 
Further improvements in the field of Object Detection (OD) were possible with networks like Regions with CNN features (R-CNN) \cite{RCNN}, Fast R-CNN \cite{fast-rcnn}, Faster R-CNN \cite{faster_r-cnn}, YOLO \cite{YOLO} and its variants. These networks and their current adaptations were the catalyst in solving some of the most challenging OD problems, enabling it to be integrated into the industry with a high degree of confidence.

\begin{figure}
    \centering
    \includegraphics[width= 0.48 \textwidth] {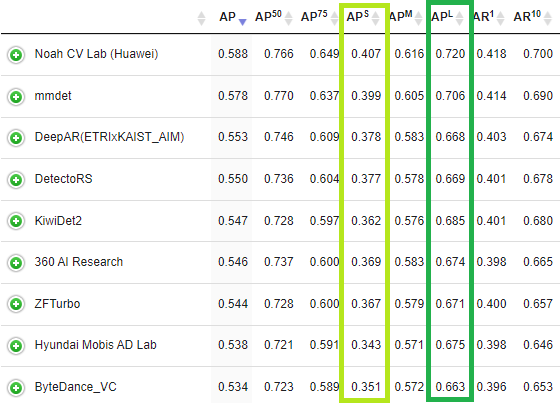}
    \caption{Detection Leader board of COCO Dataset Object Detection Challenge from 2020. The Average Precision of Small ($AP^S$) and Large ($AP^L$) objects in the dataset for the top performing models are highlighted in the figure.}
    \label{fig:OD_COCO_Leaderboard}
\end{figure}

Detecting small objects is particularly challenging as they have very small features-space to offer for training. As per COCO evaluation matrix \cite{DBLP:journals/corr/ChenFLVGDZ15/coco_eval}, the small objects are the objects in a image having an area less than 32x32 pixels. Although the state-of-the-art Machine Learning (ML) models are excellent in general OD applications involving prominently visible or dominant objects, they perform poorly in scenarios involving small objects. For instance, the latest COCO Dataset Object Detection challenge shows that even the top performing models tends to perform badly on small objects \cite{Aps_for_small} (see Figure \ref{fig:OD_COCO_Leaderboard}). More specifically, taking a YOLOv4 with CSPDarknet-53 backbone on MS COCO Dataset as example \cite{YOLOV4}, the mean average precision (mAP) for the smaller object is only 20\%, where mAP for the dominant objects are more than twice as much (mAP of 45\% for the medium and 56\% for the large objects), highlights the big gap in the performance of models in smaller OD problems.

However, when it comes to the integration of ML application into the industry, the data driven nature of these approaches make them highly dependent on the data used, both quality and quantity, which is why data availability is one of the most crucial factors in the project's success. But in industrial environments, the collection of enough useful data is unfortunately a painful process, accounting for additional costs, efforts, and time. Data has to be collected error-free and has to be categorized correctly in order to ensure its effectiveness in the ML application. Since the industrial machines of today are less prone to errors, the availability of the useful data necessary for the ML implementation becomes even more limited, turning the industrial realization a challenging and complex task. 
%
In some cases, the data obtained from the industrial environment is unbalanced, which consequently affects the model's performance on all classes. The low performance for the detection of small objects is often attributed to their unbalanced distribution over the entire dataset.
A balanced dataset could eventually improve the learning of the model and hence perform better in all associated tasks. 

As the manual labelling of small objects on image data is a time and effort intensive task, the latest advancements of synthetic data generation methods tend to address these difficulties in data preparation \cite{borkman2021unity}. For many industrial objects, CAD data is available and can be utilized as a blue-print for the generation of synthetic data. Although CAD data lacks photo-realism, a simple rendering can be imparted and the corresponding images of the target class could be generated from their CAD models without high computational costs. Upon adding the newly generated images, the class distribution of the dataset could be balanced. This approach of leveraging synthetic data generation together with real datasets for class focused data balancing and its effectiveness in improving the performance of use cases with small object detection has been addressed in this paper.


The structure of the paper is as follows: In section \ref{related_works}, we discuss the related works in OD applications in industry, as well as the various data manipulation techniques commonly used for the ML applications. This is followed by detailed description of the methodology and the experimental setup performed in section \ref{methodology}. In section \ref{results} we present the results of our approach and compares it with the other popular implementations. Finally, we conclude with the summary of the work and discussion on the future scope in section \ref{conclusion}.

\section{Background \& Related Works}\label{related_works}

Object Detection is a crucial  problem in computer vision that involves recognizing and localizing objects of interest within an image or a video stream \cite{ ren2015faster}. In recent years, deep learning (DL) techniques using CNNs, have achieved remarkable success on improving the accuracy of OD models, even in challenging  scenarios where the object features are partially occluded, poorly illuminated, or exhibit low contrast. Single Shot Detector (SSD) \cite{ liu2016ssd} and You Only Look Once (YOLO) \cite{ redmon2016you} are two CNN-based models that have gained popularity in the field of OD. These models have various applications in transportation, military,  and industrial use cases. For example, an OD-model based on YOLOv2 was utilized  in \cite{8999243} for the surface inspection on conveyor belts, while \cite{9713940} compares the effect of SSD, F-RCNN, YOLOv3 and YOLOv5 OD models in detecting surface defects in metals using a generic dataset.

Beyond general OD applications, the small OD finds its application in areal object inspection, industrial scenarios, etc. Small OD also forms the basis of many other computer vision applications such as object tracking \cite{8003302}, instance segmentation \cite{7780712}, action recognition \cite{HERATH20174} and others. The field of application of small OD is therefore diverse, where it is used for spider detection and removal application \cite{8402455}, in autonomous driving application \cite{benjumea2023yoloz}, remote sensing \cite{8517436}, or even for improving the synchronization of Digital Twin of a manufacturing system \cite{9363597}.   

Due to the data-driven nature of small OD models, they are highly dependent on the available data. In most industrial situations, collecting new datasets to train a model is a challenging, expensive and a time-consuming task \cite{challenge_article}. One way to address this is data augmentation, which is a powerful technique to alleviate over-fitting. Many researchers have proposed data augmentation methods \cite{(de_la_Rosa_2022, wang_lee_2021} to artificially increase the size of training data using the data in hand without collecting new data \cite{martins2022hybrid}. One or more morphs are applied to the data while preserving the labels during transformation \cite{salamon_bello_2017}. 

Synthetic data \cite{syndata}, and can be generated based on custom requirements and other controlled conditions. More importantly, such a dataset should follow the underlying distribution of the real dataset \cite{surv_syn}. Synthetic data generation can be achieved generally via two techniques, first by using 3D rendering and simulation tools such as Game Engines and second by using DL technique such as Variational Autoencoder \cite{VAE}, Denoising Diffusion Probabilistic Models \cite{DDPM}, GANs \cite{GANs} (such as StyleGANs3) \cite{stylegan3}, etc. 
The most simple form of synthetic data generation is using the cut-and-paste method , where one can have various types of background and the dedicated objects of different types needed to be detected. Then a combination of these objects with the background is undertaken to create an image, i.e. to cut the object and paste it on to the background image. This method is good to train OD for some simple tasks, but would not yield better results for complex OD tasks, since this approach does not help in model generalization.
While DL techniques are faster than normal cut-paste in creating a large dataset, a similar issue would arise as these models do not guarantee the desired outcome with specific angles, backgrounds, and lighting conditions. These issues can be easily tackled with synthetic data created using 3D rendering and simulation tools \cite{rajpura2017object}. 

Usage of synthetic data for OD makes many applications easier to implement. The authors, in \cite{9956710}, utilize synthetic data generated using DeepGTAV framework to work on Unmanned Aerial Vehicles scenarios. While synthetic data was leveraged in \cite{8594379} to create OD models for robotic object grasping application. CAD models offer an important asset for synthesizing image data. Leveraging the geometric precision and comprehensive annotations inherent in CAD models, it is possible to create diverse and labeled synthetic datasets that enhance the performance and generalization of OD models \cite{rajpura2017object}. Object-related CAD models are available in a highly-detailed geometric form since they are necessary attributes in manufacturing. Game engines allow the use of these CAD models of compatible formats within a simulation. Taking advantage of this feature of the renderer space and a game camera, the dataset capturing process can be simulated \cite{planche2017depthsynth}. There are also packages like Unity Perception package that enable the generation of synthetic datasets for multi-OD and segmentation applications.

In addition to the benefits associated with data generation, synthetic data generated using CAD models exhibits a notable challenge in terms of domain dissimilarity when compared to real-image-based test data.
On the other hand, the synthetic data generated from CAD models come with the problem of being inherently different to the domain of the real images. Although there are various domain adaptation techniques in the literature such as synthetic-to-real domain adaptation in \cite{Nikolenko2021}, which aim to modify synthetic data, there are also works, which present how photo-realistic rendering alone cannot reduce the domain gap and therefore attempt to resolve the issue by domain randomization \cite{tremblay2018training,jimaging8110310}. Furthermore, investigating the combined effect of synthetic and real data as a hybrid dataset, without altering its nature, presents an intriguing avenue for exploration. In scenarios where there is an amount of limited real data, this approach could help leverage it, with synthetic data alongside. 




\section{Methodology}\label{methodology}

The synthetic data generation approaches described in section \ref{related_works}, could address the data limitations hindering the successful development of small OD-models, by generating class specific balanced datasets. A photo-realistic synthetic datasets may not be necessary for some scenarios, where the generated synthetic datasets are mixed along with the available real dataset.

To validate this approach, an industrial use case of a manual assembly scenario has been taken into consideration. In the assembly process, an actor/worker performs a circuit bread-board assembly of a previously designed configuration, using the specific electronics components, while wearing an Augmented Reality (AR) or Mixed Reality (MR) glasses. The camera sensor in the AR glasses can capture the scene consisting of various objects available within the worker's field-of-view and feed it into an OD-model to perform object recognition. Based on the obtained results, a worker assistance-systems in AR will assist the worker by providing instructions and visual augmentations. The worker assistance and AR methods will not be discussed further, as they are outside the scope of this paper. The focus will be primarily only on the performance of the OD model.

After identifying the distribution of each target class in the original dataset, synthetic data generation will be used to balance the classes. With this approach, we are able to create a proportionally balanced dataset, with more instances of small objects in the combined dataset. Later, the state-of-the-art OD models were trained in combinations of experiments and the results were evaluated. 

\subsection{Data Distribution}
The data driven nature of DL/ML models makes the data distribution a significant factor in the project's success. The focus of this paper is the detection of small objects in the manual assembly scenario, where a correct identification is crucial in the worker assistance context. As the first step of this work, the original distribution of the data was identified. There were \textit{Five} objects in consideration in the target dataset, out of which the \textit{Three} objects (LED, Resister \& Button) fit the criteria of a small object i.e. below 32x32 pixels. The other \textit{Two} objects (Buzzer, Arduino) are between 32x32 pixels and 64x64 pixels and are thus considered as medium-sized objects. 

\begin{figure}
    \centering
    \includegraphics[width= 0.48\textwidth]{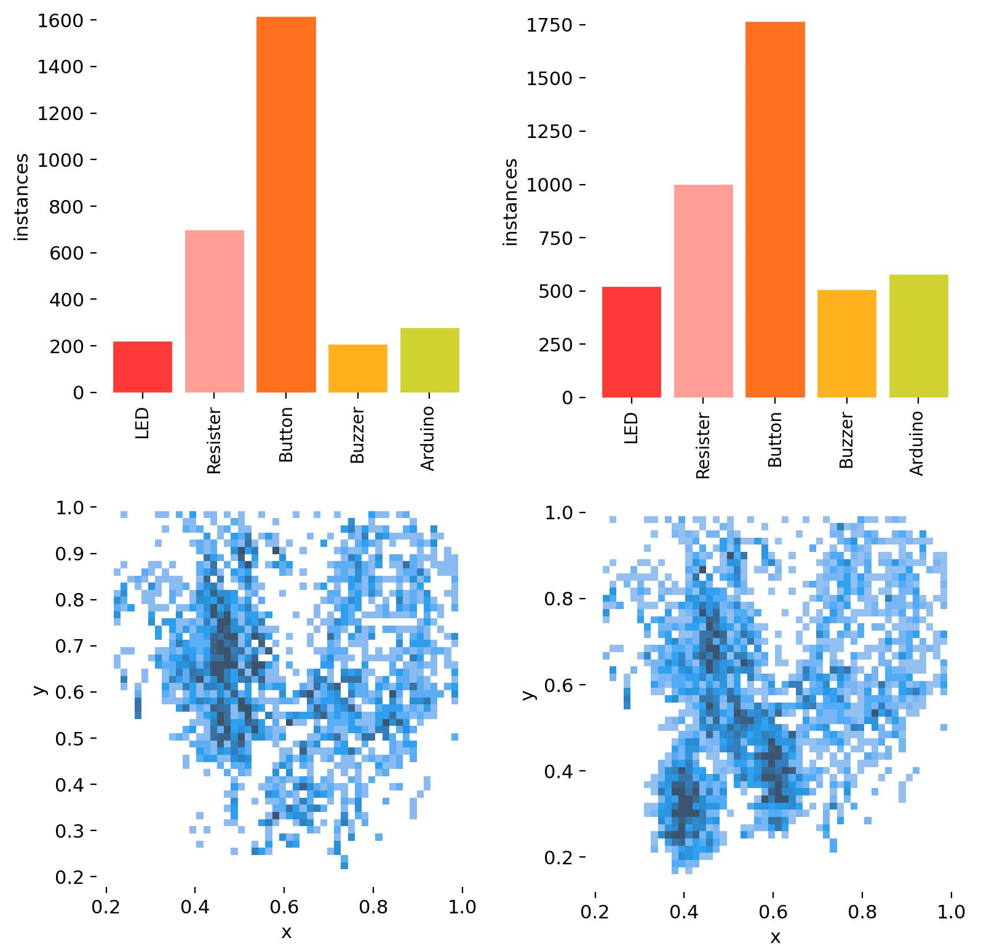}
    \caption{Data Distribution of Initial real dataset (left) to the combined dataset DS-3 (right). The target classes have been proportionally balanced, giving more significance for the less occurring classes.}
    \label{fig:data_dist}
\end{figure}

After identifying the distribution of the target objects, a proportionally balanced dataset was created using synthetic data generation. The Figure \ref{fig:data_dist} shows the initial and final distribution of the dataset (original dataset on left and the combined dataset DS-3 on the right. Details on DS-3 are shown in table \ref{tab:exp_table}).

\subsection{Dataset Generation}\label{data_generation}

In order to create a balanced dataset, synthetic data generation techniques was utilized. The table \ref{tab:exp_table} lists the datasets and their combination of real and synthetic data instances used for this work. A total of \textit{Five} datasets were generated combining the real and synthetic data and different OD models were trained on those datasets with various combinations of hyperparameters. The datasets are later referenced using the acronyms given in this table \ref{tab:exp_table}.

\begin{table}[ht]
\centering
\begin{tabular}{|c|c|c|c|c|}
\hline
Exp. No & \begin{tabular}[c]{@{}c@{}} Dataset \\Name \end{tabular} & Real Data & Synthetic Data & \begin{tabular}[c]{@{}c@{}}Total Size of  \\ Training Data\end{tabular} \\ \hline
1 & DS-1 & 300 & 0                & 300 \\ \hline
2 & DS-2 & 300 & 100 (33\% more)  & 400 \\ \hline
3 & DS-3 & 300 & 150 (50\% more)  & 450 \\ \hline
4 & DS-4 & 300 & 300 (100\% more) & 600 \\ \hline
5 & DS-5 & -   & 300              & 300 \\ \hline
\end{tabular}
\caption{The list of datasets and their division of real and synthetic data instances used. The datasets are later referenced using the acronyms given in this table.}
\label{tab:exp_table}
\end{table}

\begin{figure}
    \centering
    \includegraphics[width= 0.45\textwidth]{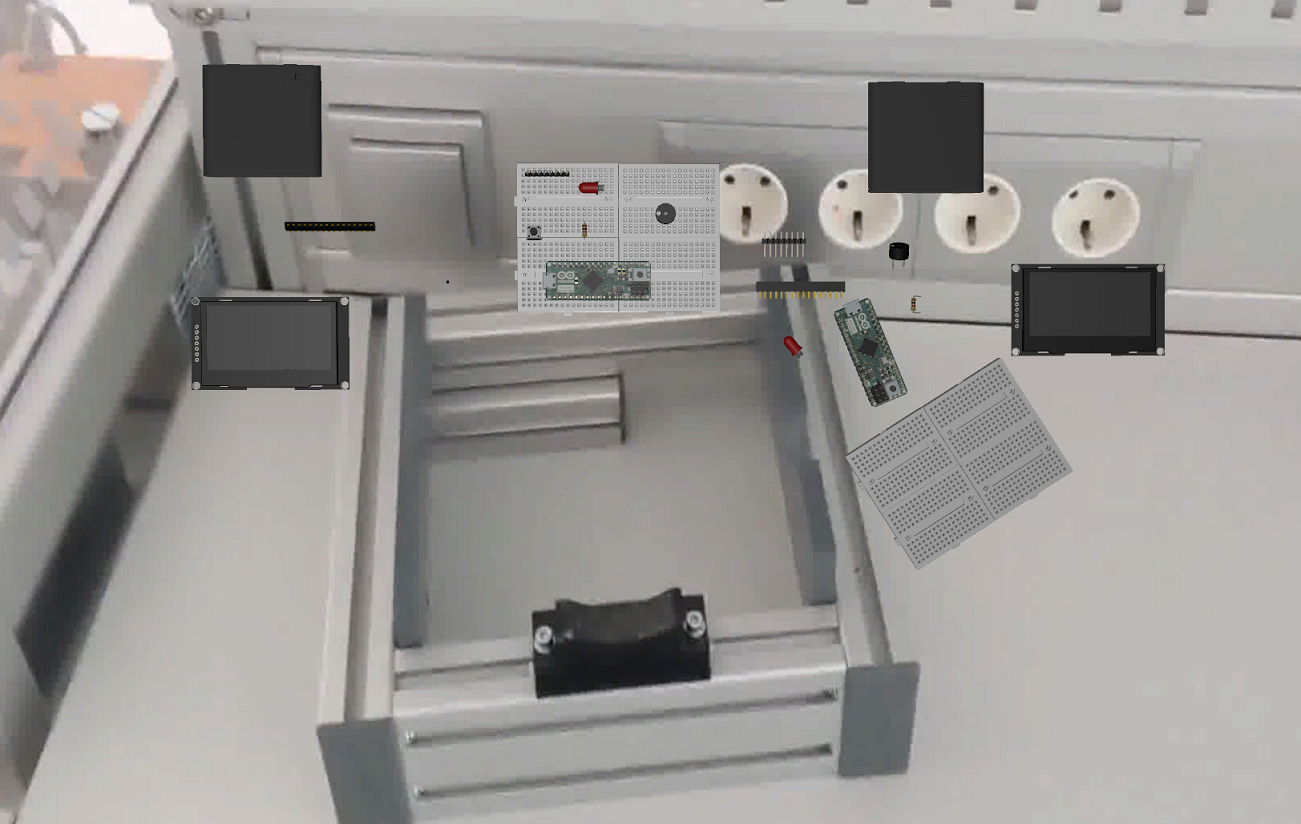}
    \caption{A sample synthetic image generated using the 3D rendering Game Engine, used in the dataset.}
    \label{fig:synthetic image}
\end{figure}

The datasets are generated using a scene simulation with a script for image capturing of the game-engine camera. Initially the CAD models are imported in the Unity scene in compatible format. On imparting the 3D models with materials and textures with minimal rendering. The dataset is simple and consists of objects in assembled and disassembled states. The object backgrounds and scales in the images are randomly set within a predefined scale range. We also varied illumination and object viewpoints. We used white lights and yellow lights with random illumination in the virtual scene while capturing images for the dataset. We do not add variable viewpoints or object occlusions in the dataset. In our use case, the objects are perceived from a limited number of viewpoints. Hence, it's a simple dataset that could be generated using the available CAD models of industrial parts. A sample of synthetic image from the dataset is shown in \ref{fig:synthetic image}.

\begin{figure}
    \centering
    \includegraphics[width= 0.45\textwidth]{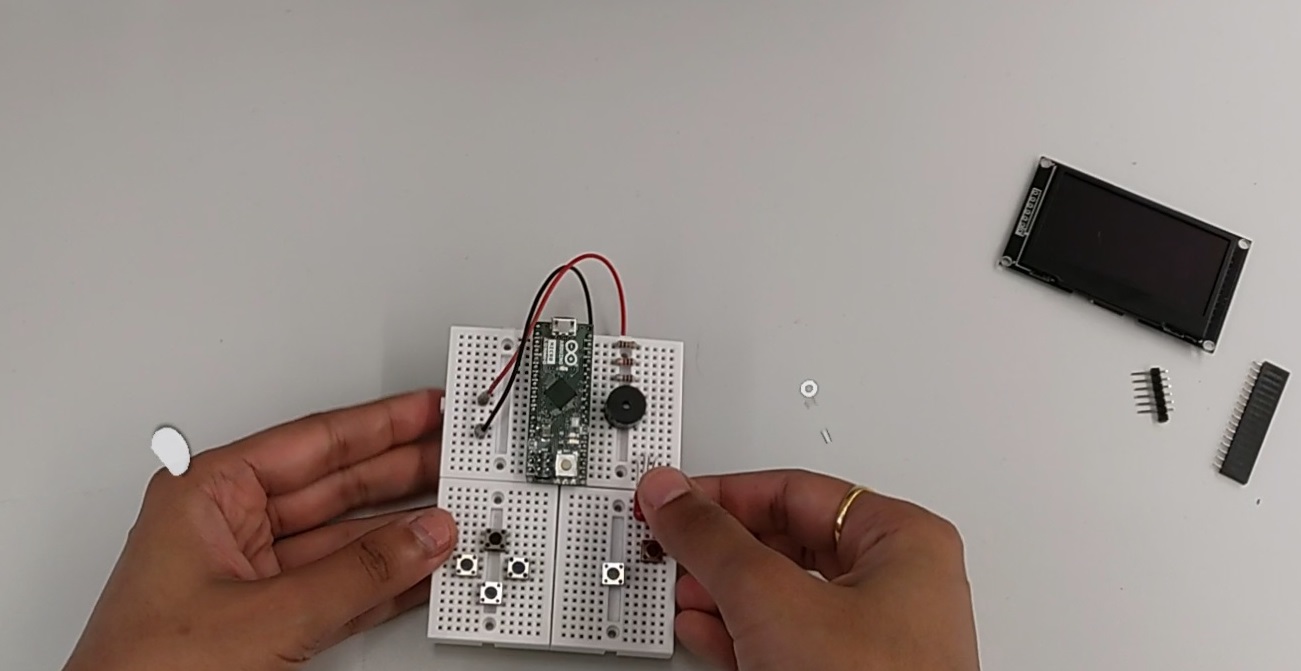}
    \caption{A sample of real image data from the Assembly Scenario, taken from the live-stream video of AR device.}
    \label{fig:Original_data}
\end{figure}

Figure \ref{fig:Original_data} shows a sample original image from the dataset and figure \ref{fig:synthetic image} shows a synthetically generated image. As it is observable, the generated data is rather simple and mainly focuses in balancing the target object instances, rather than creating a photo-realistic scene. 


\subsection{Objected Detection Models}\label{models_OD}
After generating the datasets and its combinations, they were used to train and compare state-of-the-art OD-models. Based on popularity and ease of implementation, three models were selected and utilized for the experiments. Two of the models were based on YOLO family developed on PyTorch framework and the other model was the SSD based on TensorFlow. These models are considered to be effective on small OD use cases, faster in inferences and having remarkable performances. 

\subsubsection{YOLOv5}
YOLO (You Only Look Once) is a family of OD algorithms, which performs detection in an image through a single forward pass, unlike other algorithms like Fast-RCNN or Faster-RCNN using two stages. YOLO was the first OD algorithm to combine the procedure of predicting bounding box with class label in an end to end differentiable network \cite{redmon2016look}. YOLOv5 is open source and original version of the code is written in python with PyTorch framework. YOLO models use a Non-Maxima Suppression (NMS) as a post-processing step to obtain the final bounding box for each detected object. This technique filters out redundant bounding boxes based on their overlap values and ensures that only the most relevant and accurate predictions remain in the final results \cite{YOLO}.YOLOv5 offers a range of architectures, each tailored to specific application requirements and datasets. The YOLOv5s (small), YOLOv5m (medium), YOLOv5l (large), and YOLOv5x (Extra Large) architectures exhibit increasing complexity in their design. As the architecture complexity grows, so does the accuracy of the object detection results. However, this improvement in accuracy comes at the expense of reduced model speed during object detection. Therefore, the choice of architecture depends on striking the right balance between accuracy and real-time object detection speed, considering the specific needs of the application and datasets at hand.

\subsubsection{YOLOv7}
Wang et al. in \cite{wang2023yolov7}, introduced YOLOv7 as an extension of the YOLO series of real-time OD models. It introduces various models tailored for different GPU environments, including edge GPU, normal GPU, and cloud GPU. Examples of these models include YOLOv7-tiny, YOLOv7, and YOLOv7-W6. Additionally, YOLOv7 incorporates model scaling techniques to cater to diverse service requirements, resulting in the development of models such as YOLOv7-X, YOLOv7-E6, YOLOv7-D6, and YOLOv7-E6E. They apply stack scaling to the neck component and employ the suggested compound scaling technique to increase the depth and width of the entire model, resulting in YOLOv7-X. On the other hand, for YOLOv7-W6, they utilize the newly introduced compound scaling method to derive YOLOv7-E6 and YOLOv7-D6.

YOLOv7 in OD offers notable improvements in speed and accuracy compared to previous models. The speed ranges from 5 to 160 frames per second (FPS), enabling real-time applications. In terms of accuracy, YOLOv7 achieves the highest average precision (AP) of 56.8\% among all real-time object detectors operating at 30 FPS or higher \cite{wang2023yolov7}.

\subsubsection{SSD}

SSD (Single Shot MultiBox Detector) is a prominent OD model that considers bounding box prediction as a regression problem. It starts by selecting the anchor box with the highest intersection over union (IoU) with the ground truth bounding box and gradually refines the prediction by minimizing the loss between the predicted and ground truth boxes. This iterative regression process allows SSD to achieve precise localization of objects in the image \cite{liu2016ssd}. This model provides a straightforward procedure for eliminating duplicate predictions, while SSD's regression-based approach focuses on refining the bounding box estimate by progressively minimizing the error.

\subsection{Hyper-Parameters}
After identifying and selecting the OD models as described above, the experiments have been performed. To ensure a comprehensive comparison among the three OD algorithms, it is essential to maintain consistent hyper-parameters during the training and testing phases on the identical dataset. In order to achieve this, the experiments employed the following hyper-parameters: 
\begin{itemize}
    \item Training Batch size : 8
    \item Input Image size : 1080 x 1080 pixels (height x width)
\end{itemize}

To optimize the training process, all the experiments utilized the Adam optimizer, which has been widely recognized for its efficiency in deep learning tasks. The Adam optimizer effectively combines the benefits of adaptive gradient algorithms and momentum-based optimization methods, facilitating faster convergence and improved performance during training.

To evaluate the performance of the algorithms, experiments were conducted for different numbers of Epochs: specifically,  30, 50 and 100 Epochs. By observing the algorithms' performance over multiple epochs, insights into their convergence rates and overall accuracy as the training progresses are observed.

The OD models are then trained on the dataset combinations and evaluated on a general test dataset consisted of real images. The model performances are evaluated on the COCO test evaluation matrix \cite{lin2015microsoft} and the results are reviewed and observations are noted. 

        

\begin{table*}
\centering
\begin{tabular}{|l|c|c|l|l|l|l|l|l|l|l|}
\hline
\multicolumn{1}{|c|}{No} & Epoch & Dataset               & mAP   & AP50  & APs   & APm   & APl   & ARs   & ARm   & ARl   \\ \hline
1                        & 30    & \multirow{3}{*}{DS-1} & 0.922 & 0.912 & 0.373 & 0.516 & 0.3   & 0.446 & 0.574 & 0.339 \\ \cline{1-2} \cline{4-11} 
2                        & 50    &                       & 0.946 & 0.939 & 0.47  & 0.532 & 0.35  & 0.515 & 0.588 & 0.375 \\ \cline{1-2} \cline{4-11} 
3                        & 100   &                       & 0.976 & 0.967 & 0.493 & 0.557 & 0.373 & 0.564 & 0.61  & 0.4   \\ \hline
4                        & 30    & \multirow{3}{*}{DS-2} & 0.936 & 0.929 & 0.401 & 0.511 & 0.329 & 0.476 & 0.583 & 0.364 \\ \cline{1-2} \cline{4-11} 
5                        & 50    &                       & 0.953 & 0.949 & 0.475 & 0.576 & 0.362 & 0.537 & 0.631 & 0.394 \\ \cline{1-2} \cline{4-11} 
6                        & 100   &                       & 0.955 & 0.952 & 0.543 & 0.605 & 0.366 & 0.594 & 0.654 & 0.397 \\ \hline
7                        & 30    & \multirow{3}{*}{DS-3} & 0.955 & 0.953 & 0.475 & 0.576 & 0.33  & 0.53  & 0.63  & 0.367 \\ \cline{1-2} \cline{4-11} 
8                        & 50    &                       & 0.959 & 0.953 & 0.488 & 0.566 & 0.364 & 0.354 & 0.624 & 0.398 \\ \cline{1-2} \cline{4-11} 
9                        & 100   &                       & 0.986 & 0.982 & 0.594 & 0.611 & 0.369 & 0.659 & 0.667 & 0.395 \\ \hline
10                       & 30    & \multirow{3}{*}{DS-4} & 0.966 & 0.959 & 0.449 & 0.594 & 0.346 & 0.484 & 0.661 & 0.375 \\ \cline{1-2} \cline{4-11} 
11                       & 50    &                       & 0.977 & 0.968 & 0.577 & 0.613 & 0.382 & 0.627 & 0.672 & 0.402 \\ \cline{1-2} \cline{4-11} 
12                       & 100   &                       & 0.983 & 0.979 & 0.607 & 0.628 & 0.384 & 0.658 & 0.688 & 0.41  \\ \hline
\multicolumn{1}{|c|}{13} & 30    & \multirow{3}{*}{DS-5} & 0.369 & 0.357 & 0.143 & 0.093 & 0.137 & 0.143 & 0.169 & 0.183 \\ \cline{1-2} \cline{4-11} 
14                       & 50    &                       & 0.336 & 0.323 & 0.057 & 0.104 & 0.170 & 0.057 & 0.139 & 0.207 \\ \cline{1-2} \cline{4-11} 
15                       & 100   &                       & 0.412 & 0.399 & 0.083 & 0.121 & 0.149 & 0.128 & 0.182 & 0.193 \\ \hline
\end{tabular}
\caption{Experiment Results for YOLOv5 Model, the experiment instance corresponding to DS-3 at 100 Epochs seem to give the optimum results, even comparing the DS-4 with a lot more synthetic data.}
\label{tab:results_table_Yolov5}
\end{table*}

\begin{table*}
\centering
\begin{tabular}{lccllllllll}
\hline
\multicolumn{1}{|c|}{No} & \multicolumn{1}{c|}{Epoch} & \multicolumn{1}{c|}{Dataset}               & \multicolumn{1}{l|}{mAP}   & \multicolumn{1}{l|}{AP50}  & \multicolumn{1}{l|}{APs}   & \multicolumn{1}{l|}{APm}   & \multicolumn{1}{l|}{APl}   & \multicolumn{1}{l|}{ARs}   & \multicolumn{1}{l|}{ARm}   & \multicolumn{1}{l|}{ARl}   \\ \hline
\multicolumn{1}{|l|}{1}  & \multicolumn{1}{c|}{30}    & \multicolumn{1}{c|}{\multirow{3}{*}{DS-1}} & \multicolumn{1}{l|}{0.885} & \multicolumn{1}{l|}{0.873} & \multicolumn{1}{l|}{0.259} & \multicolumn{1}{l|}{0.416} & \multicolumn{1}{l|}{0.273} & \multicolumn{1}{l|}{0.306} & \multicolumn{1}{l|}{0.487} & \multicolumn{1}{l|}{0.309} \\ \cline{1-2} \cline{4-11} 
\multicolumn{1}{|l|}{2}  & \multicolumn{1}{c|}{50}    & \multicolumn{1}{c|}{}                      & \multicolumn{1}{l|}{0.905} & \multicolumn{1}{l|}{0.897} & \multicolumn{1}{l|}{0.36}  & \multicolumn{1}{l|}{0.45}  & \multicolumn{1}{l|}{0.285} & \multicolumn{1}{l|}{0.447} & \multicolumn{1}{l|}{0.505} & \multicolumn{1}{l|}{0.328} \\ \cline{1-2} \cline{4-11} 
\multicolumn{1}{|l|}{3}  & \multicolumn{1}{c|}{100}   & \multicolumn{1}{c|}{}                      & \multicolumn{1}{l|}{0.951} & \multicolumn{1}{l|}{0.946} & \multicolumn{1}{l|}{0.48}  & \multicolumn{1}{l|}{0.566} & \multicolumn{1}{l|}{0.361} & \multicolumn{1}{l|}{0.538} & \multicolumn{1}{l|}{0.621} & \multicolumn{1}{l|}{0.393} \\ \hline
\multicolumn{1}{|l|}{4}  & \multicolumn{1}{c|}{30}    & \multicolumn{1}{c|}{\multirow{3}{*}{DS-2}} & \multicolumn{1}{l|}{0.838} & \multicolumn{1}{l|}{0.831} & \multicolumn{1}{l|}{0.331} & \multicolumn{1}{l|}{0.404} & \multicolumn{1}{l|}{0.313} & \multicolumn{1}{l|}{0.403} & \multicolumn{1}{l|}{0.461} & \multicolumn{1}{l|}{0.35}  \\ \cline{1-2} \cline{4-11} 
\multicolumn{1}{|l|}{5}  & \multicolumn{1}{c|}{50}    & \multicolumn{1}{c|}{}                      & \multicolumn{1}{l|}{0.914} & \multicolumn{1}{l|}{0.907} & \multicolumn{1}{l|}{0.463} & \multicolumn{1}{l|}{0.479} & \multicolumn{1}{l|}{0.347} & \multicolumn{1}{l|}{0.56}  & \multicolumn{1}{l|}{0.538} & \multicolumn{1}{l|}{0.372} \\ \cline{1-2} \cline{4-11} 
\multicolumn{1}{|l|}{6}  & \multicolumn{1}{c|}{100}   & \multicolumn{1}{c|}{}                      & \multicolumn{1}{l|}{0.949} & \multicolumn{1}{l|}{0.939} & \multicolumn{1}{l|}{0.499} & \multicolumn{1}{l|}{0.588} & \multicolumn{1}{l|}{0.379} & \multicolumn{1}{l|}{0.552} & \multicolumn{1}{l|}{0.635} & \multicolumn{1}{l|}{0.412} \\ \hline
\multicolumn{1}{|l|}{7}  & \multicolumn{1}{c|}{30}    & \multicolumn{1}{c|}{\multirow{3}{*}{DS-3}} & \multicolumn{1}{l|}{\textbf{0.947}} & \multicolumn{1}{l|}{\textbf{0.943}} & \multicolumn{1}{l|}{0.424} & \multicolumn{1}{l|}{0.514} & \multicolumn{1}{l|}{0.32}  & \multicolumn{1}{l|}{0.475} & \multicolumn{1}{l|}{0.588} & \multicolumn{1}{l|}{0.352} \\ \cline{1-2} \cline{4-11} 
\multicolumn{1}{|l|}{8}  & \multicolumn{1}{c|}{50}    & \multicolumn{1}{c|}{}                      & \multicolumn{1}{l|}{\textbf{0.957}} & \multicolumn{1}{l|}{\textbf{0.95}}  & \multicolumn{1}{l|}{0.42}  & \multicolumn{1}{l|}{0.575} & \multicolumn{1}{l|}{0.34}  & \multicolumn{1}{l|}{0.472} & \multicolumn{1}{l|}{0.64}  & \multicolumn{1}{l|}{0.375} \\ \cline{1-2} \cline{4-11} 
\multicolumn{1}{|l|}{9}  & \multicolumn{1}{c|}{100}   & \multicolumn{1}{c|}{}                      & \multicolumn{1}{l|}{\textbf{0.976}} & \multicolumn{1}{l|}{\textbf{0.967}} & \multicolumn{1}{l|}{0.512} & \multicolumn{1}{l|}{0.624} & \multicolumn{1}{l|}{0.38}  & \multicolumn{1}{l|}{0.553} & \multicolumn{1}{l|}{0.689} & \multicolumn{1}{l|}{0.404} \\ \hline
\multicolumn{1}{|l|}{10} & \multicolumn{1}{c|}{30}    & \multicolumn{1}{c|}{\multirow{3}{*}{DS-4}} & \multicolumn{1}{l|}{0.906} & \multicolumn{1}{l|}{0.902} & \multicolumn{1}{l|}{0.305} & \multicolumn{1}{l|}{0.538} & \multicolumn{1}{l|}{0.319} & \multicolumn{1}{l|}{0.349} & \multicolumn{1}{l|}{0.618} & \multicolumn{1}{l|}{0.354} \\ \cline{1-2} \cline{4-11} 
\multicolumn{1}{|l|}{11} & \multicolumn{1}{c|}{50}    & \multicolumn{1}{c|}{}                      & \multicolumn{1}{l|}{0.928} & \multicolumn{1}{l|}{0.922} & \multicolumn{1}{l|}{0.372} & \multicolumn{1}{l|}{0.51}  & \multicolumn{1}{l|}{0.322} & \multicolumn{1}{l|}{0.434} & \multicolumn{1}{l|}{0.616} & \multicolumn{1}{l|}{0.359} \\ \cline{1-2} \cline{4-11} 
\multicolumn{1}{|l|}{12} & \multicolumn{1}{c|}{100}   & \multicolumn{1}{c|}{}                      & \multicolumn{1}{l|}{0.946} & \multicolumn{1}{l|}{0.939} & \multicolumn{1}{l|}{0.47}  & \multicolumn{1}{l|}{0.578} & \multicolumn{1}{l|}{0.578} & \multicolumn{1}{l|}{0.519} & \multicolumn{1}{l|}{0.659} & \multicolumn{1}{l|}{0.398} \\ \hline
\multicolumn{1}{|c|}{13} & \multicolumn{1}{c|}{30}    & \multicolumn{1}{c|}{\multirow{3}{*}{DS-5}} & \multicolumn{1}{l|}{0.243} & \multicolumn{1}{l|}{0.235} & \multicolumn{1}{l|}{0.083} & \multicolumn{1}{l|}{0.059} & \multicolumn{1}{l|}{0.072} & \multicolumn{1}{l|}{0.103} & \multicolumn{1}{l|}{0.108} & \multicolumn{1}{l|}{0.147} \\ \cline{1-2} \cline{4-11} 
\multicolumn{1}{|l|}{14} & \multicolumn{1}{c|}{50}    & \multicolumn{1}{c|}{}                      & \multicolumn{1}{l|}{0.304} & \multicolumn{1}{l|}{0.3}   & \multicolumn{1}{l|}{0.105} & \multicolumn{1}{l|}{0.075} & \multicolumn{1}{l|}{0.085} & \multicolumn{1}{l|}{0.114} & \multicolumn{1}{l|}{0.117} & \multicolumn{1}{l|}{0.145} \\ \cline{1-2} \cline{4-11} 
\multicolumn{1}{|l|}{15} & \multicolumn{1}{c|}{100}   & \multicolumn{1}{c|}{}                      & \multicolumn{1}{l|}{0.374} & \multicolumn{1}{l|}{0.364} & \multicolumn{1}{l|}{0.122} & \multicolumn{1}{l|}{0.102} & \multicolumn{1}{l|}{0.129} & \multicolumn{1}{l|}{0.144} & \multicolumn{1}{l|}{0.173} & \multicolumn{1}{l|}{0.246} \\ \hline
                         & \multicolumn{1}{l}{}       & \multicolumn{1}{l}{}                       &                            &                            &                            &                            &                            &                            &                            &                            \\
                         & \multicolumn{1}{l}{}       & \multicolumn{1}{l}{}                       &                            &                            &                            &                            &                            &                            &                            &                            \\
                         & \multicolumn{1}{l}{}       & \multicolumn{1}{l}{}                       &                            &                            &                            &                            &                            &                            &                            &                           
\end{tabular}
\caption{Experiment Results for YOLOv7 Model. The DS-3 dataset clearly gives best results compared to the other dataset combinations.}
\label{tab:results_table_Yolov7}
\end{table*}

\begin{table*}
\centering
\begin{tabular}{|l|c|c|l|l|l|l|l|l|l|l|}
\hline
\multicolumn{1}{|c|}{No} & Epoch & Dataset               & mAP  & AP50& APs & APm & APl & ARs & ARm & ARl \\ \hline
1                        & 30    & \multirow{3}{*}{DS-1} & 0.492  & 0.480  & 0.015    & 0.174    & 0.254    & 0.051    &  0.251   & 0.290    \\ \cline{1-2} \cline{4-11} 
2                        & 50    &                       & 0.630  & 0.617  & 0.079    & 0.240    & 0.300    & 0.170    &  0.292   & 0.330    \\ \cline{1-2} \cline{4-11} 
3                        & 100   &                       & 0.826  & 0.765  & 0.204    & 0.329    & 0.314    & 0.265    &  0.411   & 0.347    \\ \hline
4                        & 30    & \multirow{3}{*}{DS-2} & 0.414  & 0.409  & 0.036    & 0.142    & 0.274    & 0.062    &  0.214   & 0.304    \\ \cline{1-2} \cline{4-11} 
5                        & 50    &                       & 0.556  & 0.540  & 0.079    & 0.177    & 0.263    & 0.113    &  0.232   & 0.305    \\ \cline{1-2} \cline{4-11} 
6                        & 100   &                       & 0.813  & 0.752  & 0.147    & 0.301    & 0.315    & 0.217    &  0.368   & 0.352    \\ \hline
7                        & 30    & \multirow{3}{*}{DS-3} & 0.333  & 0.307  & 0.007    & 0.087    & 0.246    & 0.027    &  0.134   & 0.289    \\ \cline{1-2} \cline{4-11} 
8                        & 50    &                       & 0.574  & 0.555  & 0.033    & 0.175    & 0.242    & 0.091    &  0.244   & 0.280    \\ \cline{1-2} \cline{4-11} 
9                        & 100   &                       & 0.850  & 0.742  & 0.190    & 0.302    & 0.329    & 0.265    &  0.379   & 0.355    \\ \hline
10                       & 30    & \multirow{3}{*}{DS-4} & 0.253  & 0.232  & 0.002    & 0.041    & 0.230    & 0.021    &  0.079   & 0.263    \\ \cline{1-2} \cline{4-11} 
11                       & 50    &                       & 0.446  & 0.430  & 0.028    & 0.144    & 0.210    & 0.072    &  0.214   & 0.242    \\ \cline{1-2} \cline{4-11} 
12                       & 100   &                       & 0.775  & 0.720  & 0.155    & 0.264    & 0.305    & 0.227    &  0.322   & 0.329    \\ \hline
\multicolumn{1}{|c|}{13} & 30    & \multirow{3}{*}{DS-5} & 0.081  & 0.076  & 0.011    & 0.034    & 0.063    & 0.062    &  0.071   & 0.092    \\ \cline{1-2} \cline{4-11} 
14                       & 50    &                       & 0.121  & 0.099  & 0.067    & 0.059    & 0.091    & 0.079    &  0.108   & 0.136    \\ \cline{1-2} \cline{4-11} 
15                       & 100   &                       & 0.198  & 0.142  & 0.088    & 0.101    & 0.123    & 0.092    &  0.132   & 0.169    \\ \hline
\end{tabular}
\caption{Experiment Results for SSD Model. SSD Model seem to perform not really good on the data, but comparing the overall performance, the model performed comparatively better on DS-3.}
\label{tab:results_table_SSD}
\end{table*}

\section{Results}\label{results}

After training the Three OD models on the five sets of datasets with three hyper-parameter combinations, the models are tested on a common test dataset, that consists of real images from the Assembly scenario. The test was carried out using the COCO evaluation matrix \cite{lin2015microsoft}, examining the performance on small, medium and large sized objects and the results are noted. Table \ref{tab:results_table_Yolov5}, Table \ref{tab:results_table_Yolov7} and Table \ref{tab:results_table_SSD} represents the evaluation results from YOLOv5, YOLOv7 and SSD models respectively. The silent feature of this table are the Average Precision and Average Recall of Small objects (APs \& ARs) which calculates the actual Precision and Recall of the model based on the true bounding box and predicted bounding box overlap for small objects (less than 32x32 pixels). This was necessary because only by observing the mAP, a true analysis cannot be made as the models are performing well on most of the cases.

Upon comparing the models, the results from Table \ref{tab:results_table_Yolov5} show that YOLOv5 performs the best out of all the three models, especially for small objects.
Upon comparing the datasets, the models trained on a dataset containing real data components (DS-1, DS-2, DS-3, DS-5) seems to perform better on the dataset with only synthetic data (DS-4). This is due to the synthetic data generation features, as mentioned in section \ref{data_generation}, where the samples created were non-photo-realistic and therefore upon testing in real dataset, the model trained only on synthetic dataset seem to perform inadequately.

It was also observed that as the number of synthetic data points increases (from DS-1 to DS-3), the models tend to learn more features through the targeted class balancing technique and reaches a maximum performance in the DS-3 dataset, where the number of synthetic data in the dataset is exactly the half of real data instances. Upon further increasing the synthetic dataset (DS-4 \& DS-5) the models tend to learn more features from the prominent synthetic data and therefore performed poorly upon testing in the real world. And comparing the hyper-parameters, the models trained for 100 epochs seem to perform better compared to the other instances. 

Comparing the recent advancements in the YOLO model families, the YOLOv7 model performs very well detecting small objects in the scenario of this work. The results of this model for different experiments which are presented in Table \ref{tab:results_table_Yolov7}, clearly show that by increasing the number of synthetic data samples, can improve the accuracy of the model to detect small objects as well as medium and large objects, but there is a limitation for that and increasing these samples too much can affect the results negatively. The results show that the highest performance occurs for DS-3, when the number of synthetic data is equal to half of real data samples.

The results of SSD model which are presented in Table \ref{tab:results_table_SSD} show that while this model can not achieve the best results compared to the other YOLO models, using and combining an amount of synthetic data with the real data samples to train the model can affect the performance and improve the accuracy of the model. The table shows that the results of training the SSD model with only synthetic data are very weak because of the difference between the synthetic dataset and the real object images, especially considering the background. On the other hand, when the amount of synthetic data samples is half of the real dataset, the performance of the SSD model is the highest (same as the YOLOv7 model), which means combining one-third of synthetic data with two-third of real dataset can affect and improve the performance of the SSD model to achieve the best result.

In a nutshell, adding synthetic data with class balancing technique can improve the performance especially for small objects, even by synthetically generating additional data points build using simple open source CAD models. The results show an increase of up to 11.4\% in precision on small objects (APs) to the base model trained only on real data set (DS-1) and best combination ratio of real and synthetic dataset (DS-3) for 100 epochs. Similar percentage increase in precision and recall can be observed for models trained on 30 and 50  epochs as well.


\section{Conclusion}\label{conclusion}

Certain scenarios, where real-world dataset are be limited and biased towards prominent objects, present challenges for detecting less prominent and significant objects, such as small-sized objects. However, by increasing the dataset through targeted class balancing techniques using synthetic data generation, there appears to be an improvement in the performance of data-driven object detection models. The results from our use case show that even with low quality of synthetic dataset, an increase of 10\% in the mAP for small objects can be achieved using this technique. But Adding more synthetic images seem to degrade the model's performance as the generated data were not photo-realistic compared to the real world data. In comparison with three state-of-the-art OD Models, YOLOv5 performed better in this particular use case. It is worth to note that, even though YOLOv7 claims to perform best for small objects \cite{wang2023yolov7}, it did not perform as well as its predecessor YOLOv5 under similar training conditions for small objects.

The task of annotating the small object dataset is tedious, and using synthetic data for automatic annotations is an optimum approach to improve the model performance. Together with class balanced data generation techniques, through synthetic data generation without implementing photo-realistic rending, a better performing models can be easily generated without much effort. For certain industrial use cases, these techniques can be really helpful in achieving the business goals.  

Although this study gives a clear comparison of state-of-the-art OD models, it has potential for many future works. By extending the quality of the generated images to more photo-realistic and bring closer to the real world could potentially improve the performance a lot. But the data diversity challenges in the real world and the corresponding model performance and comparing it to the costs of data generation can still debatable.    


\section*{Acknowledgment}
This research work is funded by the German BMBF - Bundesministerium für Bildung und Forschung (0IW19002, project InCoRAP).  We would like to thank Al Harith Farhad for proofreading the paper.





\bibliographystyle{IEEEtran}
\bibliography{IEEEabrv,biblio}

\end{document}